\documentclass[11pt,a4paper]{article}
\usepackage[hyperref]{emnlp2020}
\usepackage{times}
\usepackage{latexsym}
\usepackage{array}
\usepackage{float}
\usepackage{graphicx}

\usepackage{float}

\usepackage{microtype}

\aclfinalcopy 

\title{UIT-HSE at WNUT-2020 Task 2: Exploiting CT-BERT for Identifying COVID-19 Information on the Twitter Social Network}
\author{Khiem Vinh Tran \footnotemark \\
University of Information Technology\\
VNU-HCM, Vietnam \\
{\tt 17520634@gm.uit.edu.vn} \\
\newline \\ 
\textbf{Kiet Van Nguyen}\\
University of Information Technology\\
VNU-HCM, Vietnam \\
{\tt kietnv@uit.edu.vn} 
 \\
\\\And

\textbf{Hao Phu Phan} \footnotemark[\value{footnote}] \\ 
National Research University HSE, \\
Russia \\
{\tt ffan@edu.hse.ru} 
\newline 
\\ \\
\textbf{Ngan Luu-Thuy Nguyen}\\ 
University of Information Technology\\
VNU-HCM, Vietnam \\
{\tt ngannlt@uit.edu.vn} \\
}

\date{}

\begin{document}
\maketitle
\footnotetext{* These authors contributed equally to this work.}
\begin{abstract}

Recently, COVID-19 has affected a variety of real-life aspects of the world and has led to dreadful consequences. More and more tweets about COVID-19 has been shared publicly on Twitter. However, the plurality of those Tweets are uninformative, which is challenging to build automatic systems to detect the informative ones for useful AI applications. In this paper, we present our results at the W-NUT 2020 Shared Task 2: Identification of Informative COVID-19 English Tweets. In particular, we propose our simple but effective approach using the transformer-based models based on COVID-Twitter-BERT (CT-BERT) with different fine-tuning techniques. As a result, we achieve the F1-Score of 90.94\% with the third place on the leaderboard of 
this task which attracted 56 submitted teams in total.


\end{abstract}

\section{\textbf{Introduction}}
\label{intro}
In the mid of April 2020, the COVID-19 pandemic has caused 23M affected and more than 800,000 deaths almost over the world\footnotemark, and the number of people affected and death increase day by day. The world faces many challenges from many sides. One of the biggest problems is how to prevent the spread of  COVID-19  before the vaccine is ready. Additionally, how to provide people information about  COVID-19  with the fastest and most timely is one of the most issues. Traditionally, the information will provide some official sources as WHO or news on TV, but with the incredible speed of virus spreading, this cannot be real-time update. The social network is the channel that can solve this problem, and Twitter is one of the popular social networks with a massive amount of data. The number of tweets related to  COVID-19 on Twitter is up to 4 million daily \cite{781w-ef42-20}, and most of these tweets are uninformative. Because of that reason, detecting which tweet is informative or not is necessary. Thus, the W-NUT 2020 Shared Task 2: Identification of Informative COVID-19 English Tweets \cite{covid19tweet} attracted about 120 different teams coming from 20 countries. We present our solution for this task which is also our main contribution in this paper.
\footnotetext{https://www.worldometers.info/coronavirus/. (The COVID-19 information is updated on September 6, 2020)}

In recent years, transfer learning models such as BERT \cite{devlin2018bert}, ALBERT \cite{lan2019albert}, BERTweet \cite{nguyen2020bertweet}, XLM-R \cite{conneau2019unsupervised}, XLNet \cite{yang2019xlnet}, COVID-Twitter-BERT \cite{muller2020covid} have achieved the best performances on a variety of NLP tasks, e.g., text classification \cite{sun2019fine}, machine reading comprehension \cite{van2020new}. Fine-tuning pre-trained language models \cite{dodge2020fine} also have proved the effectiveness of exploring solutions for new tasks. In addition, ensemble approaches were also very effective in previous studies \cite{araque2017enhancing,nguyennlp,van2020job}. Inspired by these research works, we propose our simple but effective ensemble strategy using different models based on CT-BERT with different fine-tunings, achieving the F1-score of 90.94\% on the final test set. This result reaches the third rank on the leaderboard of this competition.

The rest of this paper is structured as follows. Section \ref{relate} is task definition. Section \ref{data} introduces information about the dataset .Section \ref{approach} describes our approach. Section \ref{experiment} shows experiments and results on the dataset. Finally, Section \ref{conclusion} concludes the paper and discusses future work.

\begin{figure*}[]
  \centering
  \includegraphics[width=1.0\linewidth]{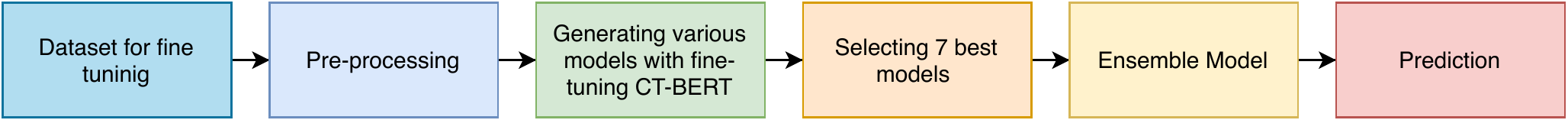}
  \caption{Our simple but effective approach for identifying COVID-19 information.}
  \label{fig:model}
\end{figure*}
\section{\textbf{Task Definition}}
\label{relate}

In this section, we summarize the W-NUT-2020 Shared Task 2 \cite{covid19tweet}.  The objective of this task is to detect whether a COVID-19 text on Twitter is informative or not.
Such informative Tweet provides information about
recovered, suspected, confirmed, and death cases
as well as the location and history of each case.
Formally, the task is described as follows.
\begin{itemize}
    \item \textbf{Input}: Given English Tweets on the social networking site Twitter.
    \item \textbf{Output}: One of two different labels (INFORMATIVE and UNINFORMATIVE) predicted by classifiers.
\end{itemize}
Several examples extracted from the dataset are presented in Table \ref{table:1}.

\begin{table}[H]
\begin{tabular}{m{6cm} c }
\hline
\multicolumn{1}{c}{\textbf{Tweet}}                                                                & \textbf{Label} \\ \hline
Oklahoma's first confirmed case of coronavirus is in Tulsa County HTTPURL \#SmartNews               & 1              \\ 
Ladies and gentlemen, put your hands together for... Johnny Covid and the Underlying Comorbidities! & 0              \\ \hline
\end{tabular}
\caption{Example of sentence and label on dataset. 0 and 1 stand for UNINFORMATIVE and INFORMATIVE, respectively.}
\label{table:1}
\end{table}





\section{Dataset}
\label{data}
We use the dataset about COVID-19 English Tweets \cite{covid19tweet} for our experiment. This corpus consists of nearly 10,000 COVID Tweets, including 4,719 Tweets annotated as informative and 5,281 Tweets annotated as uninformative. Each tweet on this dataset is annotated by three independent annotators with an inter-annotator agreement score of Fleiss' Kappa at 81.80\%. This dataset is divided into training/validation/test sets with a 70/10/20 rate.
Table \ref{tab:dataset} shows overview statistics of the dataset.

\begin{table}[H]
\centering

\resizebox{\columnwidth}{!}{\begin{tabular}{lrrrr}
\hline
\multicolumn{1}{c}{\textbf{}} &
  \multicolumn{1}{c}{\textbf{Informative}} &
  \multicolumn{1}{c}{\textbf{Uninformative}}  \\ \hline
Training &            3,303          & 3,697		\\
Validation       & 472          & 528\\ 
Test    & 944          & 1,056        \\ \hline
\end{tabular}}
\caption{Overview statistics of the dataset of WNUT-2020 Task 2.}
\label{tab:dataset}
\end{table}

\section{Our Approach}
\label{approach}
In this section, we present our simple yet effective approach for this task. Instead of leveraging multiple supper models, we only focus on the CT-BERT model for generating a variety of best-performance models by fine-turning techniques. Figure \ref{fig:model} shows the overview of the approach using three important components which are pre-processing techniques (see Subsection \ref{prepro}), the core method CT-BERT (see Subsection \ref{ctbert}), and the ensemble learning (see Subsection \ref{ensemble}). 

\subsection{Pre-processing Techniques}
\label{prepro}
Pre-processing is an essential part of the running model because it can increase the performance of the model. In this part, we try to programmatically apply some pre-processing techniques as follows: 
\begin{itemize}
\item \textbf{Step 1}: Converting Tweets into lower texts.
\item \textbf{Step 2}: Removing or replacing special characters (emojis including) with ASCII alternatives.
\item \textbf{Step 3}: Normalizing punctuation.
\item \textbf{Step 4}: BERT-tokenizer tokenizes each sentence into a list of tokens. Some kinds of contractions may be ignored because they are not listed in vocab. Thus, we use the Pycontractions tool\footnote{https://pypi.org/project/pycontractions/} with pre-trained model “glove-twitter-100” to expand the contractions.

\item \textbf{Step 5}: Expanding some common abbreviations.
\item \textbf{Step 6}: Segmenting each hashtag into words. If we feed a raw hashtag into the tokenizer, usually it will be ignored because the pre-trained model’s vocab does not contain any word like this whole hashtag. A hashtag usually contains useful information because it may mark the tweet as a newsletter. Because of that, we segment them, and their words will be less overlooked at the phase of tokenization. Segmenter tool ekphrasis \cite{baziotis-pelekis-doulkeridis:2017:SemEval2} with twitter corpus are used for this task.

\end{itemize}

\subsection{CT-BERT}
\label{ctbert}
The task is defined as the binary-classification task to detect a COVID-19 text on Twitter is informative or not. In this paper, we experiment with COVID-Twitter-BERT (CT-BERT)  \cite{muller2020covid} as our default transformer-based model. CT-BERT is based on the BERT-LARGE and is optimized to be used on the COVID-19 domain from social media. 

Following the fine-tuning pre-trained language models \cite{dodge2020fine}, we fine-tune with different parameters and random seeds. These parameters and random seeds of each model are random.  With each change of parameters and random seeds, we obtain an individual model based on CT-BERT. 

\subsection{Ensemble Learning}
\label{ensemble}
From a range of the individual models, we choose seven models with the best F1-score performances. Finally, we implement the ensemble approach using hard or soft voting to predict a final output from outputs of the seven best-performance models.

\section{Experiments}
\label{experiment}
In this section, we introduce our simple but effective ensemble model on the dataset. Our systems are evaluated using accuracy, precision (P), recall (R) and F1-score \cite{covid19tweet}. Note that P, R and F1-score are calculated on the INFORMATIVE class. 
\subsection{Data Preparation}


Because the original training set is unbalanced, we re-split this training set with a ratio of 50:50, which means that the number of informative labels and uninformative labels is approximately equal, which is more effective in the training phase.


\subsection{Experiment Setting}
We conduct various experiments on Google Colab Pro (CPU: Intel(R) Xeon(R) CPU @ 2.20GHz; RAM: 25.51 GB; GPU: Tesla P100-PCIE-16GB with CUDA 10.1). We fine-tuning CT-BERT with different parameters as batch size, learning rate, epoch, random seed. In general, we set batch size to 16 for all models. About learning rate, we set the value is 2.00E-05 for first six models and 3.00E-05 for the 7th model. The difference between each model is the epoch and random seed. As our resources are constrained, we use the method of randomly choosing random seed and try to combine them with other parameters. With the same hyper-parameter values, distinct random seeds can lead to substantially different results \cite{dodge2020fine}. After many times combine these parameters, we choose seven models with the best F1-score performances.
As a result, Table \ref{ParameterTable} shows the parameters of the seven models. 
\begin{table}[H]
\resizebox{\columnwidth}{!}{
\begin{tabular}{crrrr}
\hline
\textbf{Model} & \multicolumn{1}{c}{\textbf{Batch Size}} & \multicolumn{1}{c}{\textbf{Learning Rate}} & \multicolumn{1}{c}{\textbf{Epochs}} & \multicolumn{1}{c}{\textbf{Random Seed}} \\ \hline
1              & 16                                       & 2.00E-05                                    & 1                                    & 96                                        \\ 
2              & 16                                       & 2.00E-05                                    & 2                                    & 144                                       \\ 
3              & 16                                       & 2.00E-05                                    & 2                                    & 380,343                                    \\ 
4              & 16                                       & 2.00E-05                                    & 3                                    & 1                                         \\ 
5              & 16                                       & 2.00E-05                                    & 3                                    & 25                                        \\ 
6              & 16                                       & 2.00E-05                                    & 4                                    & 747                                       \\ 
7              & 16                                       & 3.00E-05                                    & 2                                    & 380,343                                    \\ \hline
\end{tabular}}
\caption{The parameters of the seven individual models.}
\label{ParameterTable}
\end{table}

\subsection{Experimental Results}
Table \ref{PerformanceTable} shows our experimental results. Comparing the experimental results of each model with different parameters such as epoch and random seed, we can see that Model 3 with epoch is 2 and random seed is 380,343 achieves the best accuracy of 92.41\% with F1 on validation sets. Additionally, the best precision is Model 5 with 92.36\%, and the best recall is 
Model 4 with 94.07\%. 
\begin{table}[H]
\centering
\resizebox{\columnwidth}{!}{
\begin{tabular}{lrrrrrr}
\hline
\multicolumn{1}{c}{{\bf Model}} & \multicolumn{1}{c}{{\bf P}} & \multicolumn{1}{c}{{\bf R}} & \multicolumn{1}{c}{{\bf F1}} & \multicolumn{1}{c}{{\bf Acc}} \\ \hline
1           & 0.9179    & 0.9237
        &  0.9208        & 0.9250 
                               \\ 
2            & 0.9059       & 0.9386       & 0.9220        &0.9250 
                              \\
3 & 0.9202  & 0.9280   &0.9241   & 0.9280
  \\
4           & 0.9043          & 0.9407          & 0.9221          & 0.9250  
  \\  
5           & 0.9236           & 0.9216         & 0.9226          & 0.9270
     \\
6           & 0.9076           & 0.9364         & 0.9218        & 0.9250  
   \\
7           & 0.9216         & 0.9216         & 0.9216         & 0.9260
     \\
Ensemble (SV)           & 0.9174           & 0.9407           & 0.9289
         & 0.9320
\\ 
Ensemble (HV)           & {\bf 0.9213}          & {\bf 0.9428}          & {\bf 0.9319}  
        & {\bf 0.9350}
\\\hline
\end{tabular}}
\caption{Model performances of our proposed approach on the validation set. HV, SV, P, R, F1, and Acc stand for Hard Voting, Soft Voting, Precision, Recall, F1-score and Accuracy, respectively.}
\label{PerformanceTable}
\end{table} 
However, Model 4 has the best recall, but the precision of this model is the lowest. It is a challenge for us to choose this model because precision and recall are unbalanced. The two most balanced models are Model 7 with the same precision and recall, and Model 5 with the highest precision and recall is approximately equal to the precision. Because of these challenges, we determine to use the ensemble approach with the hope of getting a better model. On this task, we use both Soft Voting (SV) and Hard Voting (HV) to ensemble these models together. As shown in Table \ref{PerformanceTable}, HV has better performance than SV when we ensemble these models together. In particular, we reach 0.9319 in F1-score on the validation set. And we achieve 0.9094 on the test set, which was ranked the third place of the leaderboard of this task.
\begin{table}[H]
\resizebox{\columnwidth}{!}{
\begin{tabular}{llrrrr}
\hline
\multicolumn{1}{c}{\textbf{Rank}} &
\multicolumn{1}{c}{\textbf{Team Name}} & \multicolumn{1}{c}{\textbf{P}} & \multicolumn{1}{c}{\textbf{R}} & \multicolumn{1}{c}{\textbf{F1}} & \multicolumn{1}{c}{\textbf{Acc}} \\ \hline
1 &NutCracker                               & 0.9135                          & 0.9057                          & 0.9096                           & 0.9150                            \\ 
2 &NLP\_North                               & 0.9029                          & 0.9163                          & 0.9096                           & 0.9140                            \\ 
3 &\textbf{UIT-HSE}               & \textbf{0.9046}                 & \textbf{0.9142}                 & \textbf{0.9094}                  & \textbf{0.9140}                   \\ 
4 &\#GCDH                                   & 0.8919                          & 0.9269                          & 0.9091                           & 0.9125                            \\ 
5 &Loner                                    & 0.8918                          & 0.9258                          & 0.9085                           & 0.9120                            \\ 
48 &Baseline                                 & 0.7730                          & 0.7288                          & 0.7503                           & 0.7710                            \\ \hline
\end{tabular}}
\caption{Performance of our system on the final scoreboard of the W-NUT-2020 Task 2. P, R, F1, and Acc stand for Precision, Recall, F1-score and Accuracy, respectively.}
\label{ScoreTable}
\end{table}

Table \ref{ScoreTable} illustrates the result of the top 5 highest teams on this shared task. As you can see from the Table \ref{ScoreTable}, with the F1 Score, our model performance is 90.94\% and just lower than the first rank team (NutCracker) and the second rank team (NLP\_North) is 0.02\% and much higher than the baseline and over 50 submitted teams.
\subsection{Error Analysis}
\begin{figure}[H]
  \centering
  \includegraphics[width=0.8\linewidth]{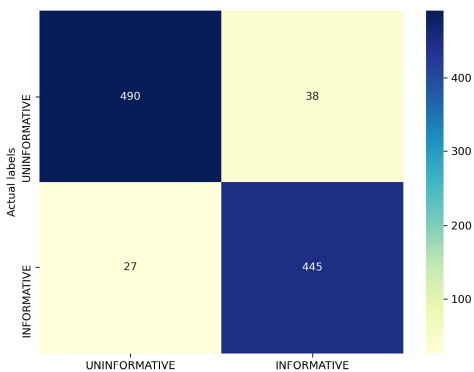}
  \caption{Confusion matrix of our best-performance model for identifying COVID-19 information.}
  \label{fig:cMatrix}
\end{figure}
Error analysis is carried out to analyze the errors that we encountered in our system by quantitative analysis using the confusion matrix of our best-performance model. Fig. \ref{fig:cMatrix} shows the confusion matrix of our best model when predicting informative tweets about COVID-19 on the validation set. It can be inferred from Fig. \ref{fig:cMatrix} that the ability of prediction of our system on the INFORMATIVE label is better than the UNINFORMATIVE label. 
As our experiment, we figure out that our wrong predicted cases are the sentences containing informative factors, but they are condition sentences or interrogative sentences. For example, Table \ref{uninformativeLabel} shows the sentences with actual labels are uninformative, but their predictions are informative.

\begin{table}[]
\resizebox{\columnwidth}{!}{
\begin{tabular}{m{0.7cm} m{10cm}}
\hline
\textbf{No.} & \multicolumn{1}{c}{\textbf{Tweet}}                                                                                                                                                                                                                                                       \\ \hline
1            & Anyone wanna tell little Amy Mexico has like 1000 Covid-19 cases to our zillions?                                                                                                                                                                                                         \\ \hline
2            & if anyone in Austin has a place to stay for just one night pls lmk I tested negative for flu, they didn't even see the need to test me for covid19. I don't feel safe going back to my apartment til I can get management involved, since my roommate owns a Glock \&amp; ran from police \\ \hline
\end{tabular}}
\caption{Examples of the sentences with actual labels are uninformative, but their predictions are informative.}
\label{uninformativeLabel}
\end{table}

\section{\textbf{Conclusion and Future Work}}
\label{conclusion}

In this paper, we propose the simple but effective ensemble approach to achieve better performance on a new dataset about COVID-19 based on tweets on Twitter and perform experiments to compare CT-BERT performances with different random seed and epoch. The experimental results showed that our proposed ensemble method with combination of data clean is effective for this dataset. Our best performance reaches 93.19\% in F1 on the validation set and 90.94\% on the test set.
Although our model performance achieves rank the third place, there is just a little gap between our model and the best model (a really small difference of 0.02\%).

In the future, we plan to combine more strong transfer learning models \cite{conneau2019unsupervised,yang2019xlnet} with different parameters to obtain higher performances. As our experiment, our ensemble model still does not reach the highest performance, so if we try with other different parameters, it will make different results. In addition, because we use max sequence length of CT-BERT is 96 tokens to fit with pre-trained data of CT-BERT, a small number of sentences that are not short enough will be truncated. We will try to analyze and handle these cases.
%



\bibliographystyle{acl_natbib}
\bibliography{W-NUT}

\begin{thebibliography}{15}
\expandafter\ifx\csname natexlab\endcsname\relax\def\natexlab#1{#1}\fi

\bibitem[{Araque et~al.(2017)Araque, Corcuera-Platas, S{\'a}nchez-Rada, and
  Iglesias}]{araque2017enhancing}
Oscar Araque, Ignacio Corcuera-Platas, J~Fernando S{\'a}nchez-Rada, and
  Carlos~A Iglesias. 2017.
\newblock Enhancing deep learning sentiment analysis with ensemble techniques
  in social applications.
\newblock \emph{Expert Systems with Applications}, 77:236--246.

\bibitem[{Baziotis et~al.(2017)Baziotis, Pelekis, and
  Doulkeridis}]{baziotis-pelekis-doulkeridis:2017:SemEval2}
Christos Baziotis, Nikos Pelekis, and Christos Doulkeridis. 2017.
\newblock Datastories at semeval-2017 task 4: Deep lstm with attention for
  message-level and topic-based sentiment analysis.
\newblock In \emph{Proceedings of the 11th International Workshop on Semantic
  Evaluation (SemEval-2017)}, pages 747--754, Vancouver, Canada. Association
  for Computational Linguistics.

\bibitem[{Conneau et~al.(2019)Conneau, Khandelwal, Goyal, Chaudhary, Wenzek,
  Guzm{\'a}n, Grave, Ott, Zettlemoyer, and Stoyanov}]{conneau2019unsupervised}
Alexis Conneau, Kartikay Khandelwal, Naman Goyal, Vishrav Chaudhary, Guillaume
  Wenzek, Francisco Guzm{\'a}n, Edouard Grave, Myle Ott, Luke Zettlemoyer, and
  Veselin Stoyanov. 2019.
\newblock Unsupervised cross-lingual representation learning at scale.
\newblock \emph{arXiv preprint arXiv:1911.02116}.

\bibitem[{Devlin et~al.(2018)Devlin, Chang, Lee, and
  Toutanova}]{devlin2018bert}
Jacob Devlin, Ming-Wei Chang, Kenton Lee, and Kristina Toutanova. 2018.
\newblock Bert: Pre-training of deep bidirectional transformers for language
  understanding.
\newblock \emph{arXiv preprint arXiv:1810.04805}.

\bibitem[{Dodge et~al.(2020)Dodge, Ilharco, Schwartz, Farhadi, Hajishirzi, and
  Smith}]{dodge2020fine}
Jesse Dodge, Gabriel Ilharco, Roy Schwartz, Ali Farhadi, Hannaneh Hajishirzi,
  and Noah Smith. 2020.
\newblock Fine-tuning pretrained language models: Weight initializations, data
  orders, and early stopping.
\newblock \emph{arXiv preprint arXiv:2002.06305}.

\bibitem[{Lamsal(2020)}]{781w-ef42-20}
Rabindra Lamsal. 2020.
\newblock \href {https://doi.org/10.21227/781w-ef42} {Coronavirus (covid-19)
  tweets dataset}.

\bibitem[{Lan et~al.(2019)Lan, Chen, Goodman, Gimpel, Sharma, and
  Soricut}]{lan2019albert}
Zhenzhong Lan, Mingda Chen, Sebastian Goodman, Kevin Gimpel, Piyush Sharma, and
  Radu Soricut. 2019.
\newblock Albert: A lite bert for self-supervised learning of language
  representations.
\newblock \emph{arXiv preprint arXiv:1909.11942}.

\bibitem[{M{\"u}ller et~al.(2020)M{\"u}ller, Salath{\'e}, and
  Kummervold}]{muller2020covid}
Martin M{\"u}ller, Marcel Salath{\'e}, and Per~E Kummervold. 2020.
\newblock Covid-twitter-bert: A natural language processing model to analyse
  covid-19 content on twitter.
\newblock \emph{arXiv preprint arXiv:2005.07503}.

\bibitem[{Nguyen et~al.(2020{\natexlab{a}})Nguyen, Vu, and
  Nguyen}]{nguyen2020bertweet}
Dat~Quoc Nguyen, Thanh Vu, and Anh~Tuan Nguyen. 2020{\natexlab{a}}.
\newblock Bertweet: A pre-trained language model for english tweets.
\newblock \emph{arXiv preprint arXiv:2005.10200}.

\bibitem[{Nguyen et~al.(2020{\natexlab{b}})Nguyen, Vu, Rahimi, Dao, Nguyen, and
  Doan}]{covid19tweet}
Dat~Quoc Nguyen, Thanh Vu, Afshin Rahimi, Mai~Hoang Dao, Linh~The Nguyen, and
  Long Doan. 2020{\natexlab{b}}.
\newblock {WNUT-2020 Task 2: Identification of Informative COVID-19 English
  Tweets}.
\newblock In \emph{Proceedings of the 6th Workshop on Noisy User-generated
  Text}.

\bibitem[{Nguyen et~al.(2019)Nguyen, Van~Nguyen, and Nguyen}]{nguyennlp}
Duc-Vu Nguyen, Kiet Van~Nguyen, and Ngan Luu-Thuy Nguyen. 2019.
\newblock Nlp@uit at vlsp 2019: A simple ensemble model for vietnamese
  dependency parsing.
\newblock \emph{The Sixth International Workshop on Vietnamese Language and
  Speech Processing VLSP 2019}.

\bibitem[{Sun et~al.(2019)Sun, Qiu, Xu, and Huang}]{sun2019fine}
Chi Sun, Xipeng Qiu, Yige Xu, and Xuanjing Huang. 2019.
\newblock How to fine-tune {BERT} for text classification?
\newblock In \emph{China National Conference on Chinese Computational
  Linguistics}, pages 194--206. Springer.

\bibitem[{Van~Huynh et~al.(2020)Van~Huynh, Van~Nguyen, Nguyen, and
  Nguyen}]{van2020job}
Tin Van~Huynh, Kiet Van~Nguyen, Ngan Luu-Thuy Nguyen, and Anh Gia-Tuan Nguyen.
  2020.
\newblock Job prediction: From deep neural network models to applications.
\newblock In \emph{2020 RIVF International Conference on Computing and
  Communication Technologies (RIVF)}, pages 1--6. IEEE.

\bibitem[{Van~Nguyen et~al.(2020)Van~Nguyen, Nguyen, Nguyen, and
  Nguyen}]{van2020new}
Kiet Van~Nguyen, Duc-Vu Nguyen, Anh Gia-Tuan Nguyen, and Ngan Luu-Thuy Nguyen.
  2020.
\newblock New vietnamese corpus for machine reading comprehension of health
  news articles.
\newblock \emph{arXiv preprint arXiv:2006.11138}.

\bibitem[{Yang et~al.(2019)Yang, Dai, Yang, Carbonell, Salakhutdinov, and
  Le}]{yang2019xlnet}
Zhilin Yang, Zihang Dai, Yiming Yang, Jaime Carbonell, Russ~R Salakhutdinov,
  and Quoc~V Le. 2019.
\newblock Xlnet: Generalized autoregressive pretraining for language
  understanding.
\newblock In \emph{Advances in neural information processing systems}, pages
  5753--5763.

\end{thebibliography}

\end{document}